# Edge Device Deployment of Multi-Tasking Network for Self-Driving Operations

Shokhrukh Miraliev
*Department of Electrical and Computer Engineering*
*Inha University*
Incheon, South Korea
shoxamiraliyev@inha.edu

Shakhboz Abdigapporov
*Department of Electrical and Computer Engineering*
*Inha University*
Incheon, South Korea
shakhboz@inha.edu

Jumabek Alikhanov
*Department of Electrical and Computer Engineering*
*Inha University*
Incheon, South Korea
jumabek4044@gmail.com

Vijay Kakani
*Department of Integrated System Engineering*
*Inha University*
Incheon, South Korea
vjkakani@inha.ac.kr

Hakil Kim
*Department of Electrical and Computer Engineering*
*Inha University*
Incheon, South Korea
hikim@inha.ac.kr

*Abstract*—A safe and robust autonomous driving system relies on accurate perception of the environment for application-oriented scenarios. This paper proposes deployment of the three most crucial tasks (i.e., object detection, drivable area segmentation and lane detection tasks) on embedded system for self-driving operations. To achieve this research objective, multi-tasking network is utilized with a simple encoder-decoder architecture. Comprehensive and extensive comparisons for two models based on different backbone networks are performed. All training experiments are performed on server while Nvidia Jetson Xavier NX is chosen as deployment device.

Keywords—self-driving, deep learning, multi-tasking network, object detection, drivable area segmentation, edge device

## I. Introduction

While recent research has been primarily focused on improving accuracy, for actual deployment in an autonomous vehicle, there are other issues of deep learning based systems that are equally critical. For autonomous driving, some basic requirements for such systems include the following: a) Accuracy of the implemented tasks. More specifically, the network ideally should achieve 100% mAP or Recall on detecting objects as well as achieving the highest mIoU possible in segmenting the drivable areas. b) Speed. The detector should have real-time or faster inference speed to reduce the latency of the vehicle control loop. c) Small model size. Smaller model size brings benefits of more efficient distributed training, less communication overhead to export new models to clients through wireless update, less energy consumption and more feasible embedded system deployment. d) Energy efficiency. Desktop and rack systems may have the luxury of burning 250W of power for neural network computation, but embedded processors targeting the automotive market must fit within a much smaller power and energy envelope.

This paper focuses on the implementation of the three most crucial tasks (i.e., object detection, drivable area segmentation and lane detection) for self-driving on embedded systems while considering the four essential requirements (i.e., model accuracy, speed, size, and energy efficiency) mentioned.

The proposed research indicates that the efficient usage of multi-task learning for merging all three tasks with one network allows better performing and energy efficient model with smaller size.

## II. Method

### A. Network architecture

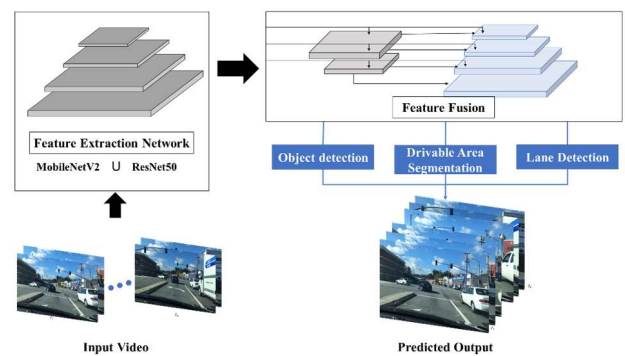

Fig. 1. Network Architecture

The model is composed of one backbone which is utilized for feature extraction from the input image, one shared feature fusion network[1] and three decoder networks for the three tasks that are handled. As a backbone, two pretrained models on ImageNet classification task are chosen for feature extraction. Since the implementation is on edge device, MobilnetV2 demonstrates faster inference speed with relatively good performance. On the other hand, Resnet50[2] backbone is larger feature extractor causing lower inference speed with higher performance. Multi-tasking model with both backbone feature extractors are deployed on an embedded system and the performance differences are compared. Secondly, feature fusion network[1] of the proposed multi-tasking model includes two parts. The first part is feature up sampling and stacking of input features

(weighted feature fusion), the second part is feature down sampling and stacking.

For Object detection, anchor based multi-scale detection scheme is adopted. Each grid of fused multi-scale features will be assigned three prior anchors with different aspect ratios and the detection head predicts the corresponding probability of each detecting category as well as the confidence of the prediction.

For the two segmentation tasks, the outputted feature maps from the feature fusion network's[1] up sampling process are restored, which represents the probability of each pixel in the input image for the drivable area segmentation/ lane detection and the background of the input. To improve the precision of the segmentation output, semantic level feature map from the backbone which is low-level feature is directly fed into the final feature fusion instead of running through the up sampling and down sampling processes.

The total loss of the multi-tasking model is the summation of the detection and segmentation losses with $\alpha$ and $\beta$ tuning parameters for a balance.

$$L_{total} = \alpha L_{det} + \beta L_{seg} \quad (1)$$

where, $L_{total}$ is a total loss, $L_{det}$ is a detection loss and $L_{seg}$ is a segmentation loss.

## B. Implementation details

The proposed multi-tasking network is trained on the Berkeley Deep Drive(BDD100K)[4] dataset which is composed of 100 thousand videos. The 10$^{th}$ second image frame of each video is annotated. The image frames are divided into 70 thousand training images and 30 thousand validation images.

Proposed multi-tasking network is trained for 200 epochs using Nvidia GeForce RTX 3090 24GB GPU. After the first 60 epochs of training the network for all three tasks, the training of segmentation heads is frozen and only object detection task is trained for another 140 epochs.

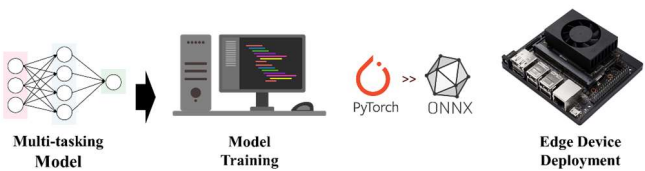

Fig. 2. The implementation process of the multi-tasking network on an embedded system

The implementation of the proposed network is done using Pytorch Framework. The trained model is converted to ONNX format for inferencing on the Nvidia Jetson Xavier NX2 edge device. The implementation process of the multi-tasking network on an edge device is shown in Fig. 2. For inferencing, batch size 1 is used and a few common resolutions are tested. The information of platforms for training and inferencing as well as the related libraries are listed in Table. 1.

TABLE I. TRAINING AND DEPLOYMENT ENVIRONMENTS

| Platform | Processor | Memory | Framework | Usage |
|---|---|---|---|---|
| Nvidia GeForce RTX 3090 | 10496 CUDA cores | 24 GB | PyTorch with CUDA 11.4 | Training Server |
| Nvidia Jetson Xavier NX2 | 384 CUDA cores | 8GB | PyTorch with CUDA 11.4 | Deployment Edge Device |

## III. EXPERIMENTS AND RESULTS

### A. Model size and running efficiency

The edge devices always have constrained the storage memories. It is necessary to consider the memory usage of each task as well as the memory usage when combining the three tasks with multi-task learning architecture. The number of parameters is highly related to the model size as well as the input resolution. At resolution 384x640, FLOPs are used to measure the computation for both models (i.e., model with Resnet50[2] and the model with MobilenetV2[3] backbones).

TABLE II. MODEL CHARACTERISTICS

| Method | # of parameters | FLOPs (384x640) | Model size |
|---|---|---|---|
| ResNet50 + Feature fusion | 27.6 M | 162.4 G | 110 MB |
| MobileNetV2 + Feature fusion | 5.4 M | 33.8 G | 24 MB |

The results indicated in Table 2. Indicates that the model with MobilenetV2[3] backbone is almost five times computationally efficient than the model with Resnet50[2] backbone, causing the multi-tasking network to have better inference speed (FPS) for real time computation. The effects of input resolution on inference speed can also be seen on Table 3. results table for models with two backbones. Model with MobileNetV2 backbone indicated almost 22 FPS inference with the lowest latency for the multi-tasking system.

TABLE III. RUNNING EFFICIENCY ON NVIDIA XAVIER NX2.

| Resolution / Model | 256x384 | 256x512 | 384x640 | 768x1280 |
|---|---|---|---|---|
| ResNet50 + Feature fusion | 68.12 ms (14.68 fps) | 85.18 ms (11.74 fps) | 139.66 ms (7.16 fps) | 473.93 ms (2.11 fps) |
| MobileNetV2 + Feature fusion | 46.15 ms (21.67 fps) | 59.71 ms (16.75 fps) | 98.23 ms (10.18 fps) | 302.11 ms (3.31 fps) |

### B. Qualitative Results

The qualitative results for the two models with different backbone networks is shown for both the highest and the lowest input resolution in Fig. 3. According to the obtained qualitative results, models with both backbones perform well in all three tasks after optimization and implementation on Nvidia Jetson Xavier NX device. Which shows that the optimization of network has minimum negative effect on network performance.

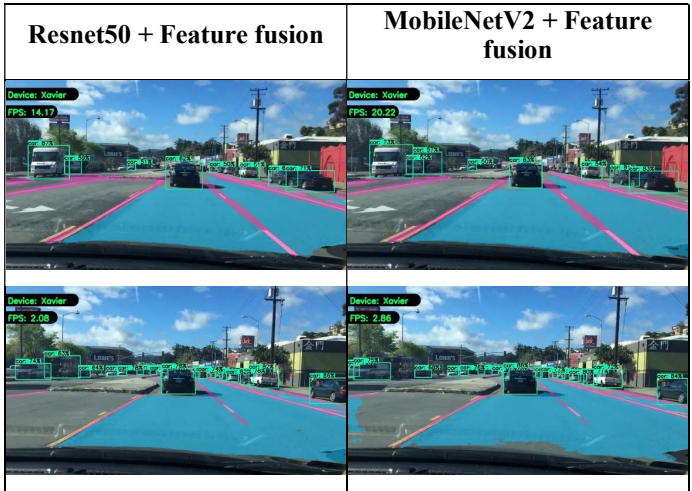

Fig. 3. Qualitative results of the implemented network with MobileNetV2[3] and Resnet50[2] backbones on Jetson Xavier NX device.

## IV. Conclusion

Running efficiency of multi-task learning models for Jetson Xavier NX device is benchmarked. Multi-tasking network is based on two popular backbones (i.e., Resnet50 and MobileNetV2). Our findings suggest that the model with MobileNetV2 backbone is only around 50% better in inference speed even though it is computationally (i.e., FLOPS) 5x times efficient than the model with Resnet50 backbone. This result implies that it is better to adapt simpler architectures based on Resnet50 for multi-tasking networks to deploy on edge devices.
ACKNOWLEDGMENT

This research was supported by the BK21 Four Program funded by the Ministry of Education(MOE, Korea) and National Research Foundation of Korea(NRF)